\documentclass[runningheads]{llncs}
\usepackage[T1]{fontenc}

\usepackage{graphicx}
\usepackage{subcaption}
\usepackage{xcolor}

\begin{document}

\title{Using Knowledge Graphs for Performance Prediction of Modular Optimization Algorithms}

\titlerunning{Using Knowledge Graphs for Algorithm Performance Prediction}

\author{Ana Kostovska\inst{1,2}\orcidID{0000-0002-5983-7169},
Diederick Vermetten\inst{4}\orcidID{0000-0003-3040-7162},
Sa\v{s}o D\v{z}eroski\inst{1,2}\orcidID{0000-0003-2363-712X},
Pan\v{c}e Panov\inst{1,2}\orcidID{0000-0002-7685-9140},
Tome Eftimov\inst{1}\orcidID{0000-0001-7330-1902}, 
Carola Doerr\inst{3}\orcidID{0000-0002-4981-3227}}

 \authorrunning{A. Kostovska et al.}

 \institute{
 Jo\v{z}ef Stefan Institute, Ljubljana, Slovenia
 \and 
 Jo\v{z}ef Stefan International Postgraduate School, Ljubljana, Slovenia
 \and 
 Sorbonne Universit\'e, CNRS, LIP6, Paris, France
 \and 
 LIACS, Leiden University, Leiden, The Netherlands
 }
% %
\maketitle              % typeset the header of the contribution
\begin{abstract}

Empirical data plays an important role in evolutionary computation research. To make better use of the available data, ontologies have been proposed in the literature to organize their storage in a structured way. However, the full potential of these formal methods to capture our domain knowledge has yet to be demonstrated.
In this work, we evaluate a performance prediction model built on top of the extension of the recently proposed OPTION ontology. 
More specifically, we first extend the OPTION ontology with the vocabulary needed to represent modular black-box optimization algorithms. Then, we use the extended OPTION ontology, to create knowledge graphs with fixed-budget performance data for two modular algorithm frameworks, modCMA, and modDE, for the 24 noiseless BBOB  benchmark functions. We build the performance prediction model using a knowledge graph embedding-based methodology. Using a number of different evaluation scenarios, we show that a triple classification approach, a fairly standard predictive modeling task in the context of knowledge graphs, can correctly predict whether a given algorithm instance will be able to achieve a certain target precision for a given problem instance. This approach requires feature representation of algorithms and problems. While the latter is already well developed, we hope that our work will motivate the community to collaborate on appropriate algorithm representations. 

\end{abstract}
\section{Introduction}

Reproducibility is slowly becoming the norm in many areas of computer science~\cite{lopez2021reproducibility}. In the domain of black-box optimization, this means that many researchers are making available not just the code, but also large amounts of benchmark data. While this increasing availability of data is beneficial to the entire community, tools to structure and interpret data are not yet widely adopted. While performance data is becoming easier to use thanks to increasing interoperability between benchmarking environments, information about the algorithms that collected that data is not as readily available. This is in part due to the complexities inherent in describing optimization heuristics. Even within a single family of algorithms, differences in operator choices, parameter adjustment strategies, and hyper-parameter settings can result in very different algorithm behavior. If these design decisions can be stored in combination with the corresponding performance data, this would open the door to extracting knowledge from the vast amount of data generated every day. One way that this could be achieved is through the use of ontologies. The data structured with the help of ontologies can then be used in various predictive studies, such as algorithm performance prediction.

In the context of computer science, \textbf{ontologies} are ``explicit formal specifications of the concepts and relations among them that can exist in a given domain''~\cite{gruberOntology}. Ontologies are generalized data models, i.e., they model only general types of things that share certain properties, but do not contain information about specific individuals in the domain. On the other hand, data about specific individuals stored in a directed labeled graph in which the labels have a well-defined meaning that comes from an ontology is commonly referred to as a \textbf{Knowledge Graph (KG)}. \textbf{Knowledge Graph Embeddings (KGEs)} are low-dimensional feature-based representations of the entities and relations in a knowledge graph. They provide a generalizable context over the entire KG, which can be used for tasks such as KG completion, triple classification, link prediction, and node classification~\cite{chen2020review,wang2017knowledge}. 

Several efforts have been made to conceptualize various aspects of domain knowledge about black-box optimization, such as Evolutionary Computation Ontology~\cite{yaman2017presenting}, Diversity-Oriented Optimization Ontology~\cite{basto2017survey}, Preference-based Multi-Objective Ontology~\cite{li2017building}, and Semantic Multi-Criteria Decision Making Ontology~\cite{mahmoudi2009semantic}. The above ontologies are strongly focused on specifics, leading to classifications of algorithms that allow users to query only for high-level relations. For example, searching for algorithms that can solve problems from a particular class, and searching for algorithms that have been applied to a specific engineering problem. The OPTimization Algorithm Benchmarking ONtology (OPTION)~\cite{kostovska2021option} formalizes knowledge about benchmarking optimization algorithms, focusing on the formal representation of data from the performance and problem landscape space, but currently lacks descriptors for optimization algorithms. 

Despite various efforts to conceptualize knowledge in black-box optimization, the main goal of these studies has been limited to the representation and organization of domain knowledge. In other words, as far as we know, the obtained representations have never been used to test their performance in predictive studies. In this work, we propose a novel approach that leverages and evaluates black-box optimization knowledge (i.e., optimization algorithms, performance data, optimization problems, and problem landscape data) represented by a formal semantic representation in the form of an ontology and KGs for predicting algorithm performance. The predictive model is developed by utilizing KG embedding-based algorithm performance classifiers.\\

\noindent\textbf{Our contributions:} We test the utility of using a formal semantic representation for black-box optimization data in the form of KGs to predict algorithm performance. The KGs contain information about the problem landscape, algorithm performance, and algorithm descriptor data. To capture this knowledge, we first extended the OPTION ontology, which already provides a vocabulary for representing problem landscape and algorithm performance data, to include representations (i.e., descriptors) that describe optimization algorithms. Proof-of-concept was performed by extending the OPTION ontology with algorithm descriptors representing two modular frameworks related to the CMA-ES (modCMA) and differential evolution (modDE) algorithms. The same ontology can be used to represent other modular algorithms. However, we note that describing all existing algorithms proposed outside of modular frameworks is a time-consuming and challenging process that requires the participation of the entire community to reach a consensus on the standard unified representation of black-box optimization algorithms.

Next, we converted the knowledge base of the extended OPTION ontology into a KG and used it to learn a binary classifier that can predict whether or not an algorithm can solve a given problem (represented as `solved' and `not-solved' relations in the KG) within a predefined target precision in a fixed-budget scenario. In the context of KGs, this task is referred to as \emph{triple classification}.

We performed an experimental evaluation of the proposed approach by predicting the performance of 324 modCMA and 576 modDE algorithm configurations on the 24 noiseless problem classes from the BBOB benchmark suite in 5 and 30 dimensions, respectively. 

We explore different evaluation scenarios to assess the predictive power of our KG-based performance classifier. The results show that our classifier correctly predicts whether an algorithm achieves a certain target precision for a given instance in the case of balanced classification. However, in the case of imbalanced classification, the baseline (the classifier that predicts the majority class) is superior.

We succeeded in improving the performance of the KG embedding-based classifier in the case of imbalanced classification by modifying the pipeline and training an additional predictive model built on the learned embeddings.
 
\noindent\textbf{Data and code availability.}
Our source code, data, the OPTION ontology extension, the generated KGs, and figures are available at: \url{https://anonymous.4open.science/r/KG4AlgorithmPerformancePrediction-63C1/;}.

\noindent\textbf{Paper outline.}
In Section~\ref{sec:formal}, we present our extension of the OPTION ontology for the formal representation of modular optimization algorithms. Section~\ref{sec:methodology} describes the construction of the KGs as well as the proposed methodology for performance prediction of modular optimization algorithms. The experimental results are discussed in Section~\ref{sec:results}. Section~\ref{sec:addressing} proposes a modification of the pipeline to address the case of imbalanced classification performance prediction. Finally, we conclude the paper with a summary of contributions and plans for future work in Section~\ref{sec:conclusions}.

\section{Formal Representation of Modular Optimization Algorithms}
\label{sec:formal}
In this paper we consider two different families of evolutionary algorithms: Differential Evolution (DE)~\cite{storn1997_de} and Covariance Matrix Adaptation Evolution Strategies (CMA-ES)~\cite{hansen1996_cmaes}. Since these two algorithms have been well-researched for over a decade, many variations and modifications have been proposed. Some of these modifications may be relatively minor, such as proposing an alternative initialization of the population. Larger changes may affect the structure of the algorithm by introducing restart mechanisms or new adaptation schemes for internal parameters. Since most of these changes are proposed in isolation, it is often difficult to understand how these variations interact. All of this has led to the development of modular algorithms. These frameworks combine large sets of variations into a single code base, where arbitrary combinations of variations can be combined into a variety of possible algorithm configurations. This not only allows a fair comparison between two different variations of the algorithm but also a more robust analysis of the potential interplay between algorithm components. 

For the CMA-ES, we use the modCMA framework~\cite{nobel_modcma_assessing}, which contains many variants of the core algorithm. This ranges from modifications of the sampling distributions (including mirrored or orthogonal sampling) to different weighting schemes for recombination to different restart strategies.

For DE, we use the modDE package available at \url{https://github.com/Dvermetten/ModDE}, v0.0.1-beta.
This framework provides a wide range of mutation mechanisms, with different modules for selecting the base component, the number of differences included, and the use of an archive for some of the different components. In addition, the usual crossover mechanisms can be enabled, as well as update mechanisms for internal parameters based on several state-of-the-art DE versions.

For the formal representation of modular optimization algorithms, we extend the OPTION ontology by creating a new ontology module that is fully compatible with OPTION. We adhere to the same ontology design principles as in OPTION. For example, we align the new classes with the same upper- and middle-level ontologies, we follow the specification-implementation-execution ontology design pattern, and we use relations from the Relations Ontology~\cite {smith2005relations}.

\begin{figure}[t]
    \centering
    \includegraphics[width=\linewidth]{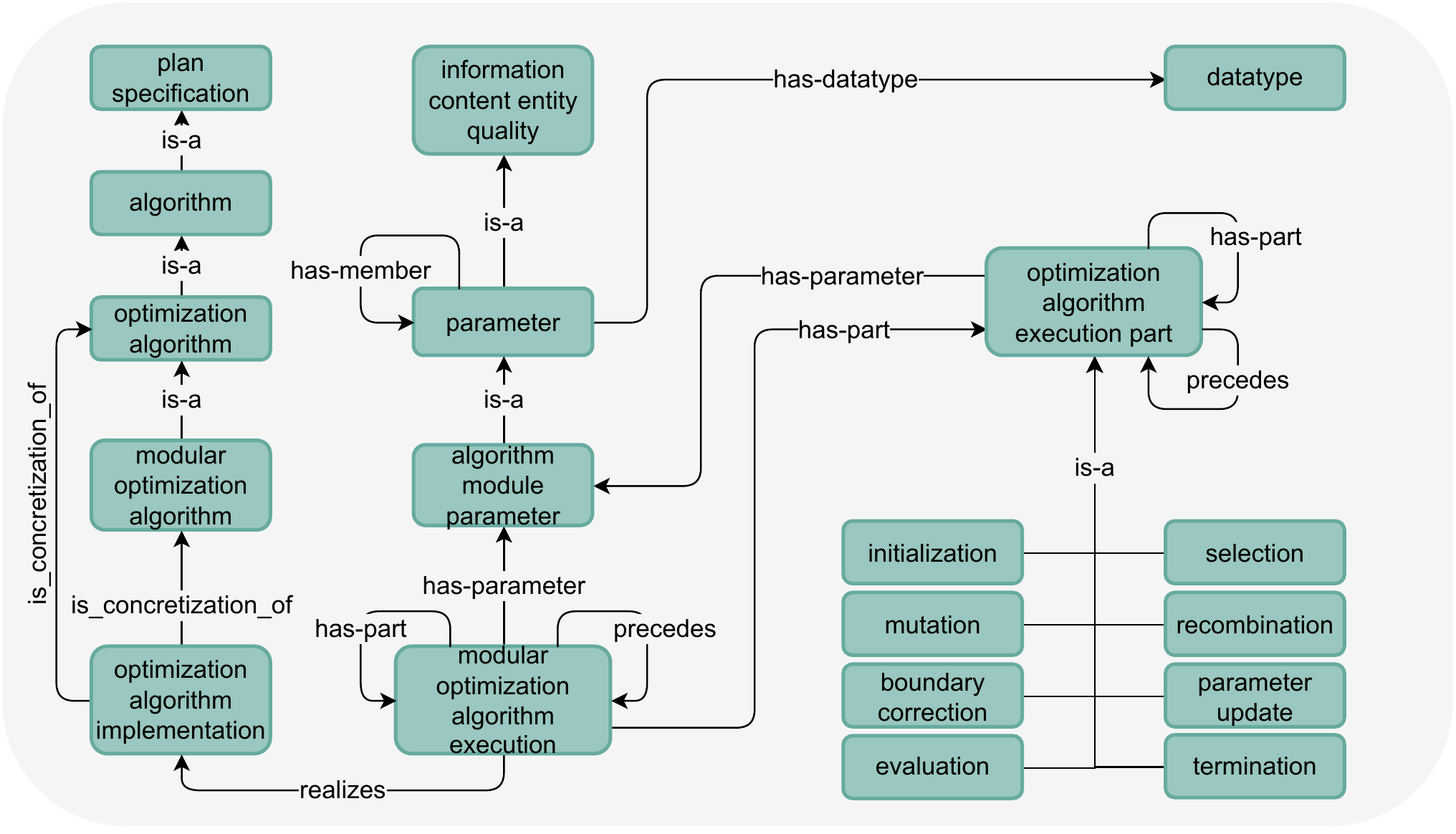}
    \caption{The entities and relations included in the extension of the OPTION ontology for the representation of modular optimization algorithms. }
    \label{fig:algo-opt}
\end{figure}

\begin{figure}
    \centering
    \includegraphics[width=\linewidth]{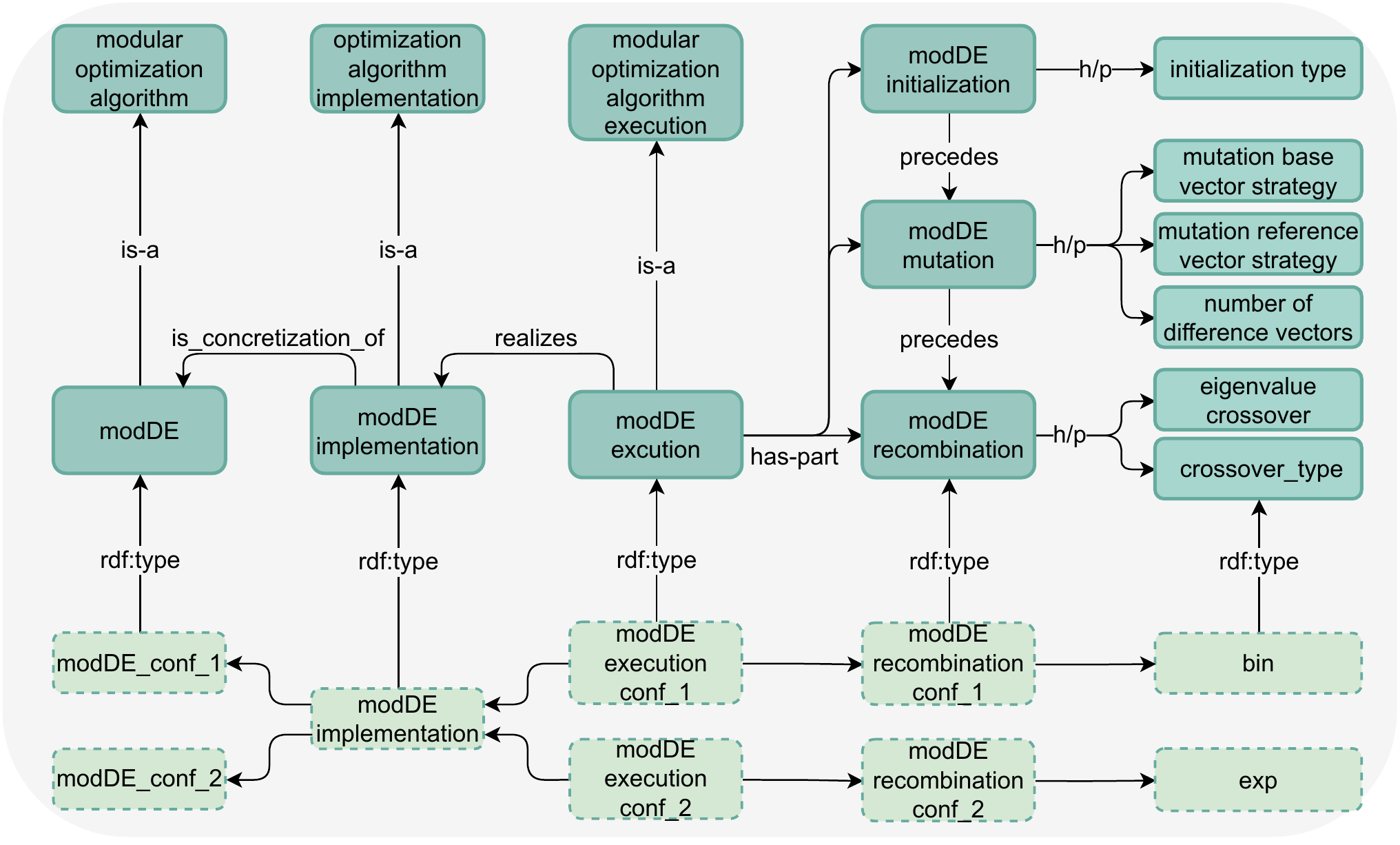}
    \caption{An illustration of the representation of the modDE algorithm in the ontology and two examples of annotation of modDE configurations. Rectangular boxes correspond to the ontology classes. Dashed rectangular boxes correspond to the class instances.}
    \label{fig:algo-opt-example}
\end{figure}

Our extension allows us to specify the different steps in the optimization process and link them to the corresponding module parameters (see Figure~\ref{fig:algo-opt}). For this purpose, we introduced the \textit{modular optimization algorithm} class~\footnote{In the rest of this paper, we will refer to the ontology classes in \textit{italic}, while the relations between the classes will be written in \texttt{typewriter}.} as a subclass of the \textit{optimization algorithm} class, which is already defined in OPTION. For modular algorithms, we have also defined a specialized class \textit{modular optimization algorithm execution}. Optimization algorithm execution can be a composition of several subprocesses (e.g., initialization, mutation, and recombination). To model this in the ontology, we have defined the \textit{modular optimization algorithm execution part} class and linked it to the \textit{modular optimization algorithm execution} class via the \texttt{has-part} relation. The algorithm execution flow is represented with the \texttt{precedes} relation. \textit{Algorithm module parameters} are linked to both \textit{modular optimization algorithm execution} and \textit{modular algorithm execution part} through the \texttt{has-parameter} relation.

In Figure~\ref{fig:algo-opt-example} we illustrate the ontological representation of the modDE algorithm. In the ontology, we create specialized subclasses of the general classes corresponding to the modDE versions. For example, the \textit{modDE execution} class is a subclass of \textit{modular optimization algorithm execution}. It inherits all the properties of its superclass but also contains definitions that are unique to the modDE algorithm, such as the different execution parts, their execution order, and links to the modDE module parameters. We note here that in Figure~\ref{fig:algo-opt-example} only the execution parts such as initialization, mutation, and recombination are shown, while the others (i.e., boundary correction, evaluation, selection, parameter update, and termination check) have been omitted due to space constraints. Finally, in Figure~\ref{fig:algo-opt-example}, we present two modDE configurations (as instances of the modDE class) that differ by the crossover type, which is a parameter that affects the recombination part of the optimization process. The modeling of the modCMA algorithm is done in a similar way.  

 The extension of the OPTION ontology, as well as the annotations of the different modCMA and modDE that we consider in this work, are available in our Zenodo repository\cite{zenodo_evo_kg_embeddings}. 

\section{Performance Prediction via KG Triple Classification}
\label{sec:methodology}
Representing optimization algorithms, benchmark problems, performance, and problem landscape data in a unified ontological framework facilitates the construction of KGs that can be used as data resources for a variety of predictive modeling tasks. In this paper, we investigate whether KG embeddings can be used to predict algorithm performance by performing the task of KG completion. More specifically, we are interested in predicting unseen performance relations between problem instances and algorithm configurations. This corresponds to the task of triple classification (i.e., whether an algorithm configuration solves a problem instance with a given target precision ). This task can be translated to a binary classification task and easily addressed by using standard machine learning algorithms. 

In this section, we first focus on the construction of the KG. Then we describe the details of the KG embedding-based pipeline for automated algorithm performance prediction. 

\begin{figure}[t]
    \centering
    \includegraphics[width=\linewidth]{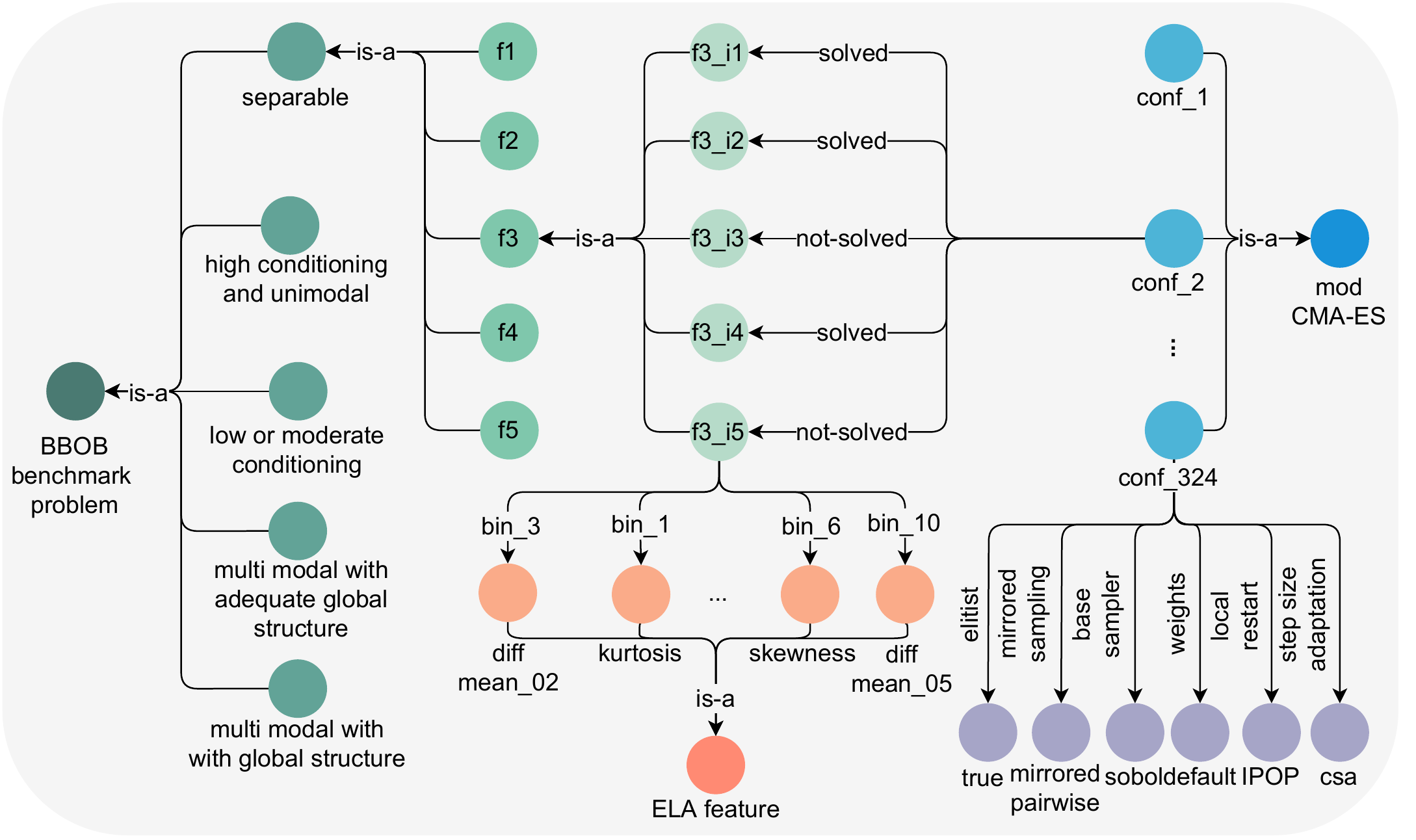}
    \caption{A snippet of our KG constructed from the original OPTION ontology and the new algorithm representation module depicting its general structure.}
    \label{fig:KG-snippet}
\end{figure}

\subsection{Construction of the KG}
The two main node types in the KGs are problem instances and algorithm instances/configurations (see Figure~\ref{fig:KG-snippet}). We collected data for the first five instances of each of the 24 noiseless BBOB problems~\cite{bbob} in dimensions $D=5$ and $D=30$, resulting in two problem sets (one for each dimension) with 120 problem instances each. Each problem instance is described with high-level and low-level landscape features. As high-level features, we used the five problem classes (i.e., separable, low or moderate conditioning, high conditioning, and unimodal multi-modal with adequate global structure and multi-modal with weak global structure) introduced in the BBOB test suite that group benchmark problems with similar properties. The low-level landscape features consist of 46 ELA features implemented in the R package \texttt{flacco}~\cite{kerschke_r-package_2016}. We use a publicly available dataset~\cite{quentin_renau_2020_3886816} containing the 46 ELA features computed for the first five instances of the 24 BBOB functions using the Sobol sampling strategy and a sample size of $100D$ with a total of 100 independent repetitions. For a more robust analysis, we use the median of the 100 calculated feature values. Finally, each ELA feature is discretized into 10 bins using the uniform binning strategy. These data are available through the OPTION ontology~\cite{kostovska2021option} knowledge base and we used their API to extract them.

The data described below were generated as part of this study and matched with data extracted from OPTION to create the KGs. The algorithm configurations are from two different modular algorithms, modular CMA-ES, and modular DE. Since it is computationally infeasible to collect data based on a complete enumeration of all possible combinations of modular CMA-ES and modular DE modules, we use a set of 324 and 576 configurations, respectively. Table~\ref{tab:confCMA} and Table~\ref{tab:confDE} contain the details of the modules and parameter spaces used, which we then use to create a Cartesian product to obtain the different algorithm configurations. In KG, each algorithm configuration is represented as a node and is connected to the different modules via labeled links/edges.

To obtain performance data for each of these configurations, we perform 10 independent algorithm runs for each problem instance and calculate the median value. As a performance measure, we use the target precision achieved by the algorithm in the context of a fixed- budget (i.e., after a fixed number of function evaluations), and use the best precision achieved after $B=\{2\,000,5\,000,10\,000,50\,000\}$ function evaluations.

\begin{table}[t]
\caption{The list of modCMA modules and their respective parameter space yielding a total of 324 algorithm configurations. }
\vspace{2ex}
\label{tab:confCMA} 
\centering
\begin{tabular}{|l|l|}
\hline
Module & Parameter space \\\hline
Elitist & True, False \\
Mirrored\_sampling &  None, mirrored, mirrored pairwise \\
base\_sampler & gaussian, sobol, halton \\
weights\_option & default, equal, 1/2\^{}lambda \\
local\_restart & None, IPOP, BIPOP \\
step\_size\_adaptation &  csa, psr\\
\hline
\end{tabular}

\end{table}

\begin{table}[t]
\caption{The list of modDE modules and their respective parameter space yielding a total of 576 algorithm configurations.}
\vspace{2ex}
\label{tab:confDE}
\centering
\begin{tabular}{|l|l|}
\hline
Module & Parameter space \\\hline
 mutation\_base & rand, best, target \\
 mutation\_reference & None, pbest, best, rand \\
 mutation\_n\_comps & 1, 2 \\
 use\_archive & True, False \\
 crossover & bin, exp \\
 adaptation\_method &  None, shade, jDE\\
 lpsr & True, False\\
\hline
\end{tabular}

\end{table}

\begin{sloppypar}
Finally, the problem instances and the algorithm configurations are associated with a \textit{solved} or \textit{not-solved} edge, depending on the performance of the algorithm considering three different target precision thresholds, $T=\{1,0.1,0.001\}$ in the case of the $5D$ benchmark problems and $T=\{10, 1,0.1\}$ for the $30D$ benchmark problems. More precisely, if an algorithm configuration achieves a target precision equal to or lower than the specified threshold for a given problem instance, we associate the algorithm configuration and the problem instance with a \textit{solved} edge; otherwise, we associate the algorithm configuration and the problem instance with a \textit{not-solved} edge.
\end{sloppypar}

\subsection{KG embedding-based pipeline for automated algorithm performance prediction}
\label{sec:KGmethodology}
Our knowledge graph $G$ can be represented as a collection of triples $\{(h,r,t)\}\subseteq E \times R \times E$, where $E$ and $R$ are the entity and relation set. One of the tasks in KG completion is to predict unseen relations $r$ between two existing entities $(h, ?, t)$. In this paper, we focus on the $\{(a, s, p)\} \subseteq A \times S \times P$ triples, where $A \subset E$ and $P \subset E$ are the algorithm configuration and the problem instance set, respectively, and $S = \{solved, not\hbox{-}solved\} \subset R$ is the performance relation. To predict the unseen performance relation between algorithm configurations and problem instances $(a, ?, p)$, we perform triple classification.

\begin{figure}[t]
    \centering
    \includegraphics[width=\linewidth]{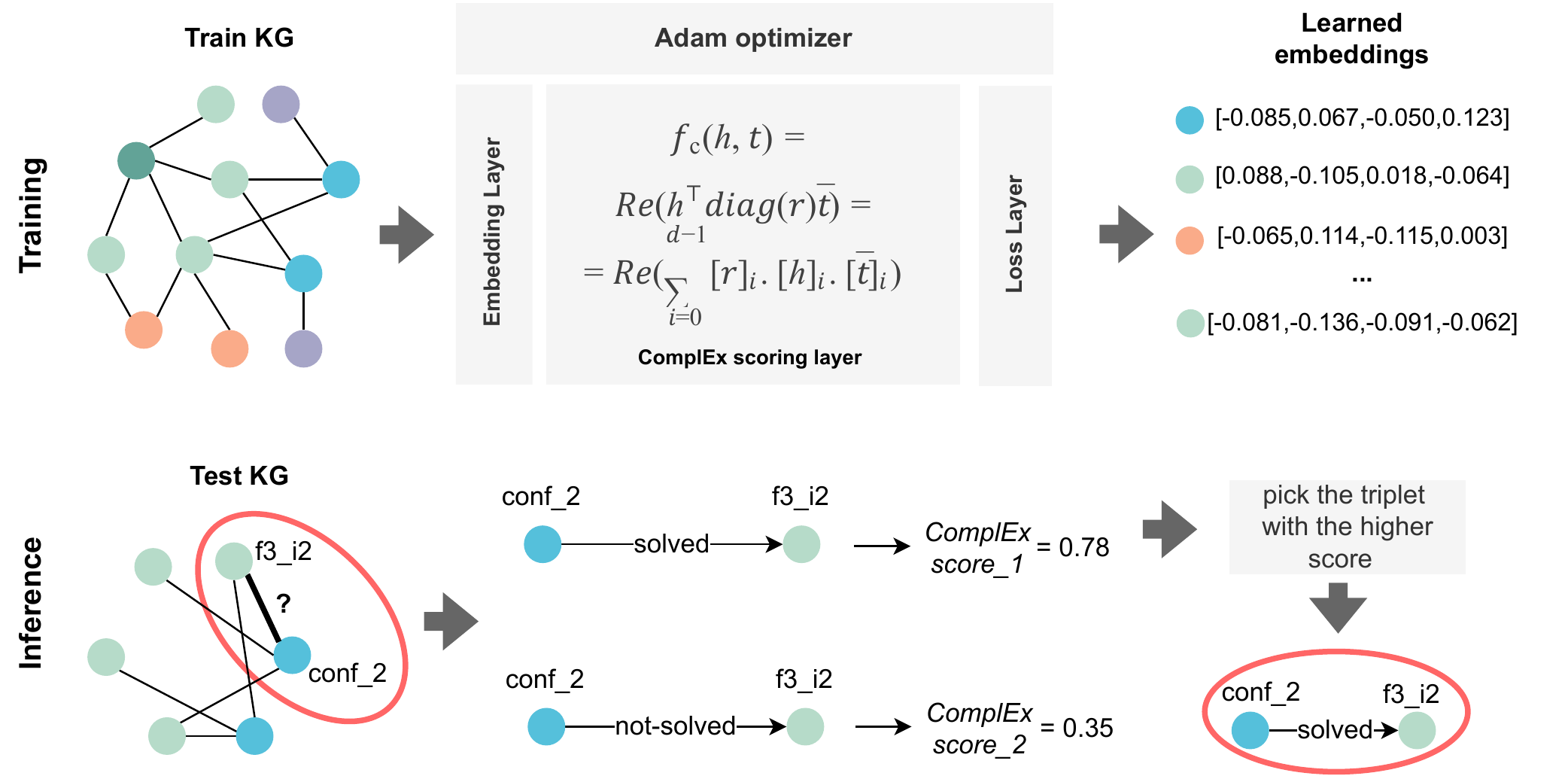}
    \caption{KG embedding-based training and inference pipeline for triple classification.}
    \label{fig:pipeline}
\end{figure}

Our proposed pipeline for predicting algorithm performance is shown in Figure~\ref{fig:pipeline}. For training the KG embeddings, we use the Ampligraph library \cite{costabello2019ampligraph}. In the training phase, we initialize the KG embeddings with the Xhavier initializer \cite{glorot2010understanding} and update them throughout several training epochs. During training, we minimize a loss function using Adam Optimizer that includes a ComplEx scoring function~\cite{trouillon2016complex} -- a model-specific function that assigns a score to a triple. Scoring functions for knowledge graph embeddings measure how far away two entities are relative to the relation in the embedding space. In general, the goal is to maximize the ComplEx model score for the positive triples and minimize it for the negative ones.

In the inference phase, we iterate over the $(a, ?, p)$ triples with a missing performance relation and calculate the ComplEx model score for the $(a, solved, p)$ and $(a, not\hbox{-}solved, p)$ triples by using the learned embeddings. We select the triple with the larger ComplEx score. 

To find the best hyperparameters for the triple classifier, we used the grid search methodology, which performs an exhaustive search over the selected hyperparameters and their corresponding search spaces. Three different hyperparameters were selected for tuning: (1) k - the dimensionality of the embedding space; (2) optimizer\_lr - the learning rate of the optimizer; and (3) loss - the type of loss function to be used during training, such as pairwise margin-based loss, negative loss probability, and adversarial sampling loss. The search spaces of the hyperparameters used in our study are shown in Table~\ref{tab:hyperparameters}.

The optimal set of hyperparameters is estimated using a separate validation set. We initially set the number of training epochs to 500, but we activate a mechanism that terminates training early if 10 consecutive validation checks/epochs do not improve performance.

For the triple classifiers, we report the F1 score, which is defined as the harmonic mean of precision and recall: $F1 = \frac{2*Precision*Recall}{Precision+Recall} = \frac{2*TP}{2*TP+FP+FN}$. 
 As a baseline, we use the classifier that predicts the majority class (solved/not-solved class). The same evaluation metric is used as a heuristic in the grid search step, where based on the F1 score we identify the best-performing model.

\begin{table}[t]
\begin{center}
\caption{Hyperparameters of the KG embedding model and their corresponding values considered in the grid search.}
\vspace{3ex}
\label{tab:hyperparameters}
\begin{tabular}{ cc }
\hline
 Hyperparameter & Search space \\ 
\hline
 k & $[50, 100, 150, 200]$\\
 optimizer\_lr & $[1e-3, 1e-4]$ \\ 
 loss & $[\textsc{pairwise},\textsc{nll},\textsc{self\_adversarial}]$ \\
 \hline
\end{tabular}
\end{center}
\end{table}

\section{Evaluation Results}
\label{sec:results}
In this section, we report the results of two different evaluation scenarios based on how we select the data for training and testing. The first one uses leave random performance triplets out validation, while the second one uses leave problem instances/algorithm configurations out validation.

\subsection{Leave random performance triplets out validation}
\label{sec:random}
For our first set of experiments, we perform algorithm performance prediction using the method described in Section~\ref{sec:methodology}. Since we consider two dimensionalities of problems, four budgets, and three target precision thresholds, we have a total of 24 different KGs for each of the two algorithms (modCMA and modDE).

For each of the KGs, we split the performance triples in the ratio 60:20:20. That is, 60\% of the triples are assigned to the training set, 20\% to the validation set, and the remaining 20\% to the test set. We do this in a stratified fashion, keeping the distribution of performance links as in the original KG. Since the split is based on a stratified sample of the performance links, performance links related to a particular problem instance or algorithm configuration can be split between the training/validation set and the test set. This approach can be used when the performance of algorithm configurations is known for the majority of problem instances in the selected problem portfolio but is unknown for some problem instances. Note that the training KG contains not only the performance triples but also other types of entities and relations, such as the high-level and low-level landscape features and the description of the algorithm configuration in terms of the modules and their parameters, while the validation and test sets contain only the links/triples of interest - `solved' and `not-solved' performance links.

The percentage of the `solved' performance relations with respect to `not-solved' ones for the modCMA and modDE KGs for the KG composite problem in 5 and 30 dimensions are shown in Table~\ref{tab:ratio}. We can notice that in some of the scenarios, we are dealing with imbalanced classification, especially in the case of $30D$ problems.

\begin{table}[!ht]
\caption{The percentage of \textit{solved} links for the modCMA and modDE algorithms in the KGs composed of a) $5D$ and b) $30D$ problems across the different fixed-budget scenarios and target precision thresholds. }
\vspace{3ex}
    \begin{subtable}[h]{\linewidth}
        \centering
        \begin{tabular}{| p{0.7cm} || c c c c || c c c c ||}
            \hline
              &  \multicolumn{4}{c||}{modCMA}  &  \multicolumn{4}{c||}{modDE} \\
              &  2000 & 5000 & 10000 & 50000 & 2000 & 5000 & 10000 & 50000 \\
            \hline
            % 10  & 0.920 & 0.947 & 0.960 & 0.980  & 0.671 & 0.787 & 0.873 & 0.961 \\
            1  & 62.9 & 68.2 & 71.3 & 78.9  & 27.7 & 42.2 & 58.1 & 81.4 \\
            0.1  & 46.8 & 54.1 & 57.1 & 63.7 & 13.2 & 23.3 & 33.1 & 62.8 \\
            0.001  & 36.9 & 47.8 & 50.7 & 55.9  & 9.4 & 14.4 & 21.7 & 56.2 \\
             \hline
        \end{tabular}
       \caption{$5D$ problems}
       \label{tab:ratio5}
    \end{subtable}
    \hfill
    \begin{subtable}[h]{\linewidth}
        \centering
        \begin{tabular}{| p{0.7cm} || c c c c || c c c c ||}
            \hline
                &  \multicolumn{4}{c||}{modCMA}  &  \multicolumn{4}{c||}{modDE} \\
                &  2000 & 5000 & 10000 & 50000 & 2000 & 5000 & 10000 & 50000 \\
            \hline
            10  & 35.1 & 46.2 & 49.9 & 68.1  & 13.0 & 21.6 & 29.1 & 46.1 \\
            1  & 10.7 & 16.0 & 21.0 & 40.9  &  1.6 & 4.4 & 8.0 & 17.5 \\
            0.1  & 6.1 & 08.8 & 12.5 & 31.5  & 1.2 & 3.2 & 6.2 & 12.7 \\
            \hline
        \end{tabular}
        \caption{$30D$ problems}
        \label{tab:accuracy}
     \end{subtable}
     
     \label{tab:ratio}
\end{table}

\begin{table}[!ht]
         \caption{The F1 score and the percentage of improvement compared to the baseline of the modCMA and modDE algorithm performance triple classifier obtained using the ComplEx scoring model for the KGs composed of $5D$ and $30D$ problems across the different fixed-budget scenarios and target precision thresholds.}
         \vspace{3ex}
    \begin{subtable}{\linewidth}
        \centering
        \begin{tabular}{| p{0.7cm} || c c c c || }
            \hline
              &  2000 & 5000 & 10000 & 50000 \\
            
            \hline
                1 & 0.922/19.43\% & 0.942/16.15\% & 0.944/13.33\% & 0.953/8.05\% \\
                0.1 & 0.905/30.22\% & 0.933/32.91\% & 0.937/29.06\% & 0.942/21.08\% \\ 
                0.001 & 0.893/15.37\% & 0.944/37.61\% & 0.944/40.48\% & 0.946/31.75\% \\ 
             \hline
        \end{tabular}
       \caption{$5D$ problems - modCMA}
       \label{tab:HPO51}
    \end{subtable}
    \begin{subtable}{\linewidth}
        \centering
        \begin{tabular}{| p{0.7cm} || c c c c || }
                \hline
              &  2000 & 5000 & 10000 & 50000 \\
              \hline
            
              1 & 0.848/1.07\% & 0.876/19.67\% & 0.901/22.59\% & 0.946/5.46\% \\
              0.1 & 0.788/-15.18\% & 0.82/-5.53\% & 0.858/6.98\% & 0.922/19.43\% \\ 
              0.001 & 0.831/-12.62\% & 0.745/-19.20\% & 0.803/-8.65\% & 0.919/27.64\% \\
             \hline
        \end{tabular}
       \caption{$5D$ problems - modDE}
       \label{tab:HPO5}
    \end{subtable}
    \hfill
    \begin{subtable}{\linewidth}
        \centering
        \begin{tabular}{| p{0.7cm} || c c c c ||}
            \hline
                &  2000 & 5000 & 10000 & 50000  \\
            \hline
            10 & 0.937/19.06\% & 0.927/32.62\% & 0.939/40.78\% & 0.953/17.51\%  \\
            1 & 0.902/-4.45\% & 0.808/-11.50\% & 0.855/-3.17\% & 0.929/25.03\%  \\ 
            0.1 & 0.935/-3.41\% & 0.89/-6.71\% & 0.852/-8.68\% & 0.921/13.28\% \\ 
            \hline
        \end{tabular}
        \caption{$30D$ problems - modCMA}
        \label{tab:HPO301}
     \end{subtable}
     \begin{subtable}{\linewidth}
        \centering
        \begin{tabular}{| p{0.7cm} || c c c c ||}
            \hline
                &   2000 & 5000 & 10000 & 50000 \\
            \hline
          
            10 &  0.9/-3.23\% & 0.931/5.92\% & 0.947/14.10\% & 0.948/35.24\% \\
            1  & 0.504/-49.19\% & 0.792/-19.02\% & 0.846/-11.69\% & 0.87/-3.76\% \\ 
            0.1 & 0.695/-30.08\% & 0.735/-25.30\% & 0.835/-13.74\%\% & 0.885/-5.04\% \\ 
            \hline
        \end{tabular}
        \caption{$30D$ problems - modDE}
        \label{tab:HPO30}
     \end{subtable}

     \label{tab:predictionHPO}
\end{table}

Table~\ref{tab:predictionHPO} presents the F1 scores of the triple classifier and the percentage of improvement of the classifier compared to the baseline across the different fixed-budget scenarios and target precision thresholds for both $5D$ and $30D$ problems. Results show that the triple classifier improves the performance, in the case when we do not have imbalanced classification. In the case of imbalanced classification, we have a performance drop, meaning that in this case our proposed pipeline should be adjusted. 

\subsection{Leave problem instances/algorithm configurations out validation}
\label{sec:prob}
Our second set of experiments evaluates a practically relevant scenario when there is no performance data for a given problem instance/algorithm configuration. To evaluate the performance of the triple classifier in this setup, we have investigated two additional evaluation scenarios:
\begin{itemize}
     \item  \textbf{Leave problem instances out validation:} In this scenario, we use all performance triples of one problem instance from each of the 24 BBOB problems for testing, select the performance triples from another problem instance for validation, and use the remaining three for training. For example, we use the first three instances of each of the 24 BBOB problems for training, the fourth instance for validation, and the fifth instance for testing. We repeat this five times so that each of the five instances appears once in the test set.
    
    \item \textbf{Leave algorithm configurations out validation}: in this scenario, the algorithm configurations are split with a 60:20:20 ratio and their performance triples are selected for training, validation, and testing, respectively. In order to assess the robustness of the results, we repeat this procedure five times independently.
    \end{itemize}

We have applied these evaluation scenarios to the KGs comprised of $5D$ benchmark problems and modCMA algorithm configurations across the four different budgets with a target precision threshold of 0.1. The average F1 scores of the triple classifier (averaged over the five runs), their standard deviations, as well as the percentage of improvement, are displayed in Table~\ref{tab:remove_problem_conf_CMA}. Similarly as in Section~\ref{sec:random}, our approach improves compared to the baseline when we have a balanced classification. 
Table~\ref{tab:remove_problem_conf_DE} presents the evaluation results for the modDE performance classifier, where similar patterns can be observed as in the previous case.

\begin{table}[t]
     \caption{The F1 score and the percentage of improvement compared to the baseline of the modCMA algorithm performance triple classifier for the KGs where all performance links are removed for a subset of problems and algorithm configurations composed of 5D problems across the different budgets and a target precision threshold of 0.1.}
         \vspace{3ex}
    \centering
    \begin{tabular}{|r|| c | c |}
    \hline
   
                & Leave-problems-out  & Leave-algorithms-out \\
                 \hline
         2000 & 0.728 (0.006)/4.90 & 0.893 (0.009)/28.67 \\
         5000 & 0.761 (0.018)/8.40 & 0.915 (0.011)/30.34 \\
         10000 & 0.766(0.008)/5.36 & 0.91 (0.011)/25.17 \\
         50000 & 0.797(0.014)/2.44 & 0.913 (0.002)/17.35 \\
          \hline
    
    \end{tabular}

    \label{tab:remove_problem_conf_CMA}
\end{table}

\begin{table}[t]
    \caption{The F1 score and the percentage of improvement compared to the baseline of the modDE algorithm performance triple classifier for the KGs where all performance links are removed for a subset of problems and algorithm configurations composed of 5D problems across the different budgets and a target precision threshold of 0.1.}
         \vspace{3ex}
    \centering
        \begin{tabular}{|r|| c | c |}
    \hline
   
                & Leave-problems-out  & Leave-algorithms-out \\
                 \hline
         2000 &  0.854(0.061)/-8.07\% & 0.79(0.035)/-14.96\% \\
         5000 &  0.837(0.022)/-3.57\% & 0.85(0.021)/-2.07\% \\
         10000 & 0.796(0.024)/-0.75\% & 0.825(0.010)/2.87\%\\
         50000 & 0.83(0.010)/7.51\% & 0.822(0.013)/6.48\% \\
          \hline

    \end{tabular}

    \label{tab:remove_problem_conf_DE}
\end{table}

\section{Addressing the problem of imbalanced classification}
\label{sec:addressing}
To solve the problem that arises in the case of imbalanced classification, we modify the pipeline described in Section~\ref{sec:KGmethodology}. More specifically, after the KG training phase, we add an additional training layer, where we train a Random Forest (RF) classifier based on the learned embeddings. Our data instances are the performance triples. In order to generate the data for the RF classifier, we represent each $(a, s, p)$ triple as a concatenation of the embedding vectors of the $a$ and $p$ entities. 
 
We perform inference by using the RF classifier instead of the ComplEx model scores. 

We evaluate this approach using the most imbalanced scenario from the experiments in Section~\ref{sec:random}, i.e., the setup where we predict modDE performance on $30D$ problem instances with a target precision threshold of 0.1. We train RF classifier with 10 estimators, implemented in the scikit-learn library\cite{scikit-learn}. The rest of the hyperparameters are used with their default values. In Table~\ref{tab:smote}, we compare the F1 scores of the classifiers trained using the pipeline presented in Section~\ref{sec:KGmethodology} with the scores of the RF classifiers described in this section. The results show that training a RF classifier on the learned embeddings improves the performance in terms of F1 score. As we are dealing with imbalanced classification, the choice of the evaluation measure is essential. Here, we also report AUC ROC score as a second evaluation measure. We reach the same conclusion with the AUC ROC evaluation measure, as the RF classifier achieves AUC ROC scores of 0.994, 0.998, 0.998, 0.987 for the 2000, 5000, 10000 and 50000 fixed budget scenarios, respectively. Our baseline (classifier predicting the majority class) has an AUC ROC score of 0.5 across the different fixed-budget scenarios. We believe that the results improve because there is a separability in the embeddings space that the RF models manage to capture when predicting the algorithm's performance. However, this assumption requires further investigation. 

\begin{table}[t]
\caption{Comparison of the two proposed pipelines for modDE performance prediction on the $30D$ problem instances with a target precision of 0.1. Results are reported in the format: F1-score of the classifier/F1-score of the baseline/Percentage of improvement compared to the baseline.}
\vspace{3ex}
\centering
        \begin{tabular}{| p{1cm} || c | c | c||}
            \hline
                % &   \multicolumn{4}{c||}{modDE} \\
                &   KG - ComplEx scoring  & RF classifier   \\
            \hline
                2000 & 0.695/0.994/-30.08\% &  0.999/0.994/0.52\% \\ 
                5000 & 0.735/0.984/-25.30\% &  0.998/0.984/1.43\%  \\ 
                10000 & 0.835/0.968/-13.74\% & 0.996/0.968/2.91\% \\ 
                50000 & 0.885/0.932/-5.04\% & 0.991/0.932/6.27\% \\ 
            \hline
        \end{tabular}
    
    \label{tab:smote}
\end{table}

\section{Conclusions and Future Work}
\label{sec:conclusions}
In this paper, we investigate the predictive power of a formal semantic representation of black-box optimization for automated prediction of algorithm performance. To this end, we evaluate the feasibility of using KGs to predict the performance of the modCMA and modDE optimization algorithms on the noiseless BBOB functions. More specifically, our goal was to investigate whether we can train KG embeddings that can be used to predict performance links/triplets (\textit{solved} or \textit{not-solved} links) in the KG between algorithm configurations and problem instances with a given target precision in a fixed- budget scenario. The KGs combine the problem landscape and algorithm performance data with the data related to the modular algorithm configuration.

The results show that when we randomly select performance triples for the test set (a classic KG completion scenario), our proposed triple classifier outperforms the baseline in the cases where we have balanced classification. In the case of imbalanced classification, the performance of the classifier decreases and it is worse than the baseline. In our second set of experiments, we have a ''more rigorous'' evaluation scenario where we try to predict all the performance links belonging to the problem instances and algorithm configurations that appear in the test set (no performance links appear in the training set). We observe similar patterns as in the previous case. To solve the performance degradation problem in the case of imbalanced classification, we modify the proposed pipeline and train a Random Forest classifier on top of the learned embeddings. 

For our future work, we plan to test different methods for training KG embeddings and improve their explainability. It would also be interesting to compare our results with other approaches such as relational learning and graph frequent pattern mining. 

We have shown that KGs of experimental data about the modCMA and modDE modular optimization algorithms created from ontology knowledge bases can be used in predictive studies. It would be interesting to test the applicability of this approach to other modular frameworks. However, the more challenging task would be to extend it outside of modular frameworks, as we would need to develop a formal, standard vocabulary that can be used to represent algorithm operators, their hyperparameters, and interactions. We hope that our work will inspire the community to collaborate on the development of appropriate algorithm representations.

\vspace{0.7ex}
\small{
 \textbf{Acknowledgments.} 
The authors acknowledge the support of the Slovenian Research Agency through program grant No. P2-0103 and P2-0098, project grants N2-0239 and J2-4460, a young researcher grant to AK, and a bilateral project between Slovenia and France grant No. BI-FR/23-24-PROTEUS-001 (PR-12040), as well as the EC through grant No. 952215 (TAILOR). Our work is also supported by ANR-22-ERCS-0003-01 project VARIATION, and via a SPECIES scholarship for Ana Kostovska.}
\bibliographystyle{splncs04}
\bibliography{mybibliography}

\end{document}